\documentclass{Interspeech2024}
\usepackage{graphicx}
\usepackage{latexsym}
\interspeechcameraready

\usepackage{soul}
\usepackage{url}
\usepackage{graphicx}
\usepackage{amsmath}
\usepackage{amsthm}
\usepackage{booktabs}
\usepackage{algorithm}
\usepackage{algorithmic}

\usepackage{colortbl,booktabs}
\usepackage{threeparttable}
\usepackage{tabularx}
\usepackage{multirow}
\usepackage{multicol}
\usepackage{mathtools}
\usepackage{amsmath,amsfonts} 
\usepackage{subfigure}
\usepackage{bm}
\usepackage{xcolor}
\usepackage{color}
\usepackage{newfloat}
\usepackage{listings}
\usepackage{caption}

\usepackage{times}
\usepackage{latexsym}

\usepackage[T1]{fontenc}

\usepackage[utf8]{inputenc}

\usepackage{microtype}
\usepackage{amsfonts}
\usepackage{amssymb}
\usepackage{multirow}
\usepackage{adjustbox}
\usepackage[english]{babel} 
\usepackage{arydshln}
\usepackage{color}
\usepackage{booktabs}
\usepackage{CJKutf8}
\usepackage{color}
\usepackage{amsmath}
\usepackage{graphicx}
\usepackage{tikz}
\usepackage{hyperref}

\title{Large Language Models for Expansion of Spoken Language Understanding Systems to New Languages}


\name[affiliation=*{1,2}]{Jakub}{Hoscilowicz}
\name[affiliation={1}]{Pawel}{Pawlowski}
\name[affiliation={1}]{Marcin}{Skorupa}
\name[affiliation={2}]{Marcin}{Sowa\'{n}ski}
\name[affiliation={2}]{Artur}{Janicki}

\address{
  $^1$Samsung R\&D Institute Poland, Warsaw, Poland\\
  $^2$Warsaw University of Technology, Poland
}

\email{\thanks{*correspondence to {\scriptsize\texttt{\textless jakub.hoscilowicz.dokt@pw.edu.pl\textgreater}}}}
\begin{document}

\maketitle

\begin{abstract}

Spoken Language Understanding (SLU) models are a core component of voice assistants (VA), such as Alexa, Bixby, and Google Assistant. In this paper, we introduce a pipeline designed to extend SLU systems to new languages, utilizing Large Language Models (LLMs) that we fine-tune for machine translation of slot-annotated SLU training data. Our approach improved on the MultiATIS++ benchmark, a primary multi-language SLU dataset, in the cloud scenario using an mBERT model. Specifically, we saw an improvement in the Overall Accuracy metric: from \SI{53}{\percent} to \SI{62.18}{\percent}, compared to the existing state-of-the-art method, Fine and Coarse-grained Multi-Task Learning Framework (FC-MTLF). In the on-device scenario (tiny and not pretrained SLU), our method improved the Overall Accuracy from \SI{5.31}{\percent} to \SI{22.06}{\percent} over the baseline Global-Local Contrastive Learning Framework (GL-CLeF) method. Contrary to both FC-MTLF and GL-CLeF, our LLM-based machine translation does not require changes in the production architecture of SLU. Additionally, our pipeline is slot-type independent: it does not require any slot definitions or examples. Code and info on model checkpoint are available at \href{https://github.com/Samsung/MT-LLM-NLU}{\color{blue}{https://github.com/Samsung/MT-LLM-NLU}}.
\end{abstract}

\section{Introduction}

The Spoken Language Understanding (SLU) module is a core component of mobile voice assistants (VAs), and it primarily focuses on two main tasks: determining the user's intent and filling in the relevant information slots, e.g., extracting named entities. The development of SLU models is heavily reliant on extensive labeled datasets. This requirement presents a significant barrier to achieving high-performance SLU models for low-resource languages that lack such comprehensive resources.

Current approaches~\cite{cheng2023fc, qin2022gl, brown2020language} to cross-lingual SLU often resort to code-switching methods to overcome the language barrier. While these methods have shown promising performance, they are not without their limitations, especially when dealing with low-resource languages.

On the other hand, machine translation (MT) of SLU training data has faced challenges due to difficulties in slot alignment between languages. However, this paper presents a breakthrough showcasing LLMs' superior ability in handling the slot translation issue. Our LLM-based MT method represents a significant advancement in overcoming the obstacles faced by conventional MT approaches in the context of cross-lingual SLUs, e.g.,~\cite{xu2020end}.

The main aspect of our approach is adopting the EasyProject~\cite{chen2023easyproject} method for transfer slot annotation. In this method, the named entities are annotated with HTML-like tags.

The effectiveness of our approach is validated through comprehensive experiments, where we achieve state-of-the-art performance on the MultiATIS++ dataset (main multilingual SLU benchmark). We substantiate the high quality of our translated data by successfully training a small on-device model from scratch (not-pretrained SLU), achieving an average improvement of \SI{17}{\percent} relative compared to GL-CLeF~\cite{qin2022gl}.

As a result, our proposed methodology not only achieves state-of-the-art performance on the MultiATIS++ dataset but also presents a paradigm shift in addressing multilingual SLU challenges. By demonstrating the superiority of LLM-based MT, we showcase a streamlined, scalable approach that outperforms complex alternatives and does not require changes in the SLU production architecture. Due to the fact that our approach is slot-agnostic (does not require any slot descriptions or examples), it can be easily employed in various production scenarios. Our contributions pave the way for enhanced multilingual understanding, making our method a promising solution for real-world applications.

\section{Related Work}

In the field of MT for SLU in VAs, advancements have been made in addressing the complex task of translating and aligning slots in multilingual contexts. This section reviews notable contributions in this evolving domain.

Historically, slot transfer in translation has relied on unsupervised word alignment from statistical MT~\cite{schuster2019cross} or attention weights from neural MT models~\cite{garg2019jointly}. Recent developments, such as the joint architectures by Xu et al.~\cite{xu2020end} and the ``Translate-and-Fill'' method by Nicosia et al.~\cite{nicosia2021translate}, have focused on translating and re-annotating slots in target sentences.

Pre-LLM MT models, key in VA advancements, have been employed by researchers to enhance slot transfer capabilities, building upon earlier MT developments. De Bruyn et al.~\cite{de2022machine} employed the NLLB model, a derivative of M2M100, for training a multilingual intent classification (IC) and slot-filling (SF) model, using data from the MASSIVE~\cite{fitzgerald2022massive} and SLURP~\cite{bastianelli2020slurp} datasets, as well as GPT-3 generated content. In contrast to their approach, our research aims to develop a broadly applicable MT model for VAs, inclusive of diverse data from various VA domains and designed to be neutral regarding intent and slot types. Following this, Sowanski et al.~\cite{sowanski2023slot} introduced an M2M100-based MT model for VAs, capable of transferring SLU slots between languages, demonstrating a \SI{17.21} BLEU points improvement in the VA domain over the baseline M2M100 model.

In Global-Local Contrastive Learning Framework (GL-CLEF) \cite{qin2022gl}, the authors introduced a framework aimed at enhancing zero-shot cross-lingual SLU by leveraging contrastive learning to bring closer representations of similar sentences across languages. By utilizing bilingual dictionaries to create multilingual code-switched data pairs, GL-CLEF effectively aligns sentence meanings across languages, improving the model's ability to understand and process multilingual information without direct translation.

Brown et al.~\cite{brown2020language} explored the capabilities of pre-trained LLMs in direct translation tasks, showing their effectiveness without specific translation training. Rosenbaum et al.~\cite{Rosenbaum2022linguist} presented the LINGUIST model, leveraging the AlexaTM 5B LLM for enhanced performance in IC and SF tasks, with a +\SI{1.9} point increase in IC Recall and a +\SI{2.5} point rise in SF F1 Score on the SNIPS dataset. Hoang et al.~\cite{hoang2023fly} demonstrated an ensemble method, combining MT models with a 7 billion parameter LLM to boost translation quality. Their approach achieved a \SI{0.6} COMET improvement in de-en translations, surpassing the MT model despite the LLM's lower score of \SI{0.9} COMET.  

The Fine- and Coarse-
grained Multi-Task Learning Framework (FC-MTLF)~\cite{cheng2023fc} employs a multi-task learning strategy by introducing an auxiliary multilingual neural machine translation (NMT) task to mitigate the limitations of code-switching. Additionally, the curriculum learning strategy is adopted to enhance performance further. Experimental results indicate that the FC-MTLF framework is current state-of-the-art on the MultiATIS++ dataset. 

In contrast to the methods and models mentioned in this section, our data-based approach does not depend on predefined slot types or any specific SLU architecture. 

\begin{table*}[hbt!]
\caption{On-Device scenario results for 5MB ROM SLU models on the MultiATIS++ dataset, comparing Not-Pretrained Transformer + LLM Slot Translator (NPT + LLM-ST, ours) with BiLSTM + GL-CLeF. Overall Accuracy was used to measure the performance of models.} \label{table:XLMRoberta_Effectiveness}
\centering
    \begin{tabular}{l|cccccccc|c}
            \hline

            \textbf{Overall Accuracy} & \textbf{de} & \textbf{es} & \textbf{fr} & \textbf{hi} & \textbf{ja} & \textbf{pt} & \textbf{tr} & \textbf{zh} & \textbf{AVG} \\
            \hline
            
            BiLSTM + $\textsc{GL-CLeF}$ \cite{qin2022gl}  & 4.60 & 9.10 & 4.30 & 0.34 & 2.03 & 16.82 & 2.80 & 2.46 & 5.31\\
        

            NPT + LLM-ST (ours) & \textbf{26.99} & \textbf{27.21} & \textbf{8.29} & \textbf{10.64} & \textbf{7.61} & \textbf{29.56} & \textbf{23.50} & \textbf{42.67} & \textbf{22.06} \\
            
            \hline
    \end{tabular}
\end{table*}

\section{Method}

The proposed method initiates with human-labeled SLU data in English. This data is translated by an LLM into multiple languages: German (de), Spanish (es), French (fr), Hindi (hi), Japanese (ja), Portuguese (pt), Turkish (tr), and Chinese (zh). The translated datasets are then utilized to train SLU models. 

\subsection{Slot Transfer Task}

The Slot Transfer Task is the core challenge for a cross-lingual SLU. It involves accurately annotating named entities during the translation process. This task is key for keeping the MT-based multilingual SLU system's performance consistent, requiring the correct transfer of slot types and their values. For example, transferring the \textit{music\_artist} slot from English to another language means the slot value, like ``radiohead'', stays the same, while its annotation is correctly placed in the translated sentence. 

Previously, authors often used an additional model specifically for slot alignment. However, with the development of deeper MT models, as discussed by De Bruyn et al.~\cite{de2022machine}, the translation model itself can directly handle this task. The introduction of LLMs has further improved the translation and slot transfer quality, streamlining the process and enhancing the overall effectiveness of multilingual SLU systems.

\subsection{Machine Translation}

In this study, we leverage the capabilities of MT, specifically through the fine-tuning of the BigTranslate LLM~\cite{BigTranslate}, to address the task of Named Entity Slot Translation. The cornerstone of our methodology is the EasyProject approach~\cite{chen2023easyproject}, which involves the utilization of parallel translation pairs annotated with named entity slots. These slots are marked within the sentences using HTML-like tags to demarcate entities, allowing for a structured approach to translation that retains the semantic integrity of named entities across languages.

Central to our methodology is employing the MASSIVE parallel open-source dataset, which serves as a critical resource for fine-tuning LLM. This dataset, known for its extensive compilation of parallel sentence pairs in various languages, was instrumental in our study. We modified annotations within the dataset with HTML-like tags, using the idea from EasyProject, e.g., <a>, <b>. For instance, an English sentence:

\vspace{6pt}

\noindent \textit{My name is <a> John <a> and I live in <b> London <b>.}
\vspace{6pt}

\noindent is matched with its German translation:
\vspace{6pt}

\noindent \textit{Mein Name ist <a> John <a> und ich wohne in <b> London <b>''.}

This deliberate annotation of named entities in both the source and target texts is essential for training the BigTranslate LLM, as it enables the model to accurately learn the complexities of translating text while maintaining the integrity of named entity slots.

\begin{table*}[hbt!]
	\centering
	\caption{Cloud scenario results for the MultiATIS++ dataset, utilizing an mBERT-based SLU system. We compare the performance of various methods, including ours (BigTranslate + LLM Slot Translator, abbreviated as LLM-ST) against others. The table presents Overall Accuracy, the most important metric from the production perspective.}\label{table:result_oa}
    \begin{tabular}{l|cccccccc|c}
        \hline
        \textbf{Overall Accuracy}  & \textbf{de} & \textbf{es} & \textbf{fr} & \textbf{hi} & \textbf{ja} & \textbf{pt} & \textbf{tr} & \textbf{zh} & \textbf{AVG} \\ 
        \hline
        AR-S2S-PTR \cite{rongali2020don}                        & 34.00 & 40.72 & 17.22 & 7.45 & 10.04 & 33.38 & -- & 23.74 & -\\
        IT-S2S-PTR* \cite{zhu2020don}                             & 39.46 & 50.06 & 46.78 & 11.42 & 12.60 & 39.30 & -- & 28.72 & -\\
        mBERT  \cite{devlin2018bert}                      & 52.69 & 52.02 & 37.29 & 4.92 & 7.11 & 43.49 & 4.33 & 18.58 & 27.25 \\
        CoSDA \cite{chen2019bert}                  & 57.06 & 46.62 & 50.06 & 26.20 & 28.89 & 48.77 & 15.24 & 46.36 & 39.90 \\
        \textsc{GL-CLeF} \cite{qin2022gl}                         & 66.03 & 59.53 & 57.02 & 34.83 & 41.42 & 60.43 & 28.95 & 50.62 & 49.85 \\
        \textsc{FC-MTLF} \cite{cheng2023fc}                        & 69.54 & 61.43 & 59.62 & 36.86 & 44.64 & \textbf{64.55} & 30.86 & 56.52 & 53.00 \\
        \hline
    BigTranslate (zero-shot) & 64.73 & 37.74 & 53.42 & 32.59 & 0.90 & 53.30 & 0.56 & 50.06 & 36.66 \\
    BigTranslate + LLM-ST (ours)  & \textbf{72.45} & \textbf{65.17} & \textbf{63.83} & \textbf{58.45} & \textbf{52.41} & 60.69 & \textbf{54.55} & \textbf{69.88} & \textbf{62.18} \\

        \hline
    \end{tabular}
\end{table*}

\subsection{Joint SLU} \label{jointnludesc}
In this paper, we focus on SLU within the context of VAs, specifically addressing the critical tasks of IC and SF~\cite{weld2022survey}. An intent represents a command uttered to a dialogue system and slots are parameters of that command. For example, in the utterance ``play Radiohead on Spotify'', `Radiohead' and `Spotify' are slots and `to\_play\_music' is an intent.

\
As our SLU architecture, we used the framework outlined by Chen et al.~\cite{chen2019bert}, utilizing multilingual BERT (mBERT) as a core pre-trained model. The architecture features two separate softmax classifiers, one for intent detection and another for SF. To assess the quality of the SLU system, we utilized semantic frame accuracy, referred to as `Overall Accuracy'. It measures the percentage of test sentences where the system correctly predicts both the intent and all slots. It is the most important metric from a production perspective; for a given user command, both intent and all slots must be predicted correctly to trigger the proper action on the device.

We utilized only the English training set from the MultiATIS++. This dataset, renowned for its detailed annotations of intents and slots across a spectrum of dialogues in the air travel domain, served as a foundational resource. To demonstrate the efficacy of LLM in translating SLU datasets, we subsequently translated the English train set into several other languages. This translation process was designed to validate the capability of our LLM to produce datasets that could effectively train SLU models across various linguistic contexts.

Following the translation of the English train set, we rigorously tested our SLU model using the original MultiATIS++ test sets in the respective languages. This testing phase was critical for evaluating our methodology's effectiveness, particularly in assessing the quality of the translated datasets. By directly comparing the performance of our SLU model on these original test sets, we could ascertain the translation's fidelity and its impact on the model's learning and generalization capabilities.

Leveraging the high-quality datasets generated through our LLM's translation process, we initiated the training of a compact, on-device SLU model from scratch (Not-Pretrained Transformer). The on-device model, trained on our translated datasets, exhibited an average improvement of over \SI{17}{\percent} relative on the MultiATIS++, compared to the GL-CLEF method applied to not-pretrained neural network of comparable size (BiLSTM).

\section{Experiments}

In our experiments, we evaluated our method across two widely used scenarios: cloud-based and on-device SLU. Each scenario offers distinct challenges and advantages. Cloud-based SLU systems have access to significant computational resources, allowing for larger models that help achieve multilingual capabilities. Conversely, on-device SLU models are limited by the hardware capacities of devices like smartphones or TVs. They focus on fast execution and privacy but their small size, necessary to fit within device memory, presents obstacles to supporting multilingual functionality. By testing in both settings, our study shows the effectiveness of our method, BigTranslate + LLM Slot Translator (LLM-ST), in advancing multilingualism under the different conditions of cloud-based and on-device SLU systems.

\begin{table*}[hbt!]
	\centering%
	\caption{Cloud scenario results for the MultiATIS++ dataset, utilizing an mBERT-based SLU system. The table presents SF and IC accuracy compared to LINGUIST (as they did not report results for Overall Accuracy).} \label{table:result}
    \begin{tabular}{l|cccccc|cc}
        \hline
        \textbf{Intent Accuracy}  & \textbf{de} & \textbf{es} & \textbf{fr} & \textbf{hi} & \textbf{ja} & \textbf{pt} & \textbf{AVG} \\ 
        \hline
       LINGUIST \cite{Rosenbaum2022linguist} & 94.08 & 97.10 & 96.88 & 94.08 & \textbf{95.38} & 92.86 &  95.06 \\
       \hline
        BigTranslate (zero-shot) & \textbf{96.86} & 97.20 & 97.54 & 94.29 & 5.49 & \textbf{97.98}  & 81.56 \\
       BigTranslate + LLM-ST (ours) & 95.52 & \textbf{98.21} & \textbf{98.32} & \textbf{95.41} & 94.06 & 97.76 & \textbf{96.54} \\
       \hline
       \hline
        \textbf{Slot F1}  & \textbf{de} & \textbf{es} & \textbf{fr} & \textbf{hi} & \textbf{ja} & \textbf{pt} & \textbf{AVG} \\ 
        \hline
       LINGUIST & 84.61 & 86.89 & 83.83 & 76.61 & \textbf{86.32} & 85.63 & 83.98 \\
       \hline
       BigTranslate (zero-shot) & 84.88 & 55.39 & 79.63 & 58.73 & 57.18 & 77.96 & 68.96 \\
       BigTranslate + LLM-ST (ours) & \textbf{92.34} & \textbf{89.10} & \textbf{87.43} & \textbf{84.35} & 81.11 & \textbf{85.72} & \textbf{86.68} \\
        \hline
    \end{tabular}
\end{table*}

\subsection{Experimental Setup}

Our study leveraged the 13B parameter BigTranslate LLM to investigate MT-LLMs' efficacy in the context of multilingual SLU. We decided to go with BigTranslate for our experiments because, unlike many other LLMs that mainly focus on a few languages, BigTranslate is a specialized version of the LLaMA model that has been fine-tuned specifically for machine translation. It can handle over 100 languages, making it an excellent tool for our study, especially since we were looking at diverse languages (including low-resource languages like Turkish). The experimental framework was structured around the MASSIVE~\cite{fitzgerald2022massive} dataset, chosen for its extensive coverage of parallel slot-annotated translation pairs across 51 languages.

The MASSIVE dataset was divided according to its original train-validation split, maintaining the integrity of the dataset's structure for training purposes. Notably, the ``annot\_utt'' field within the MASSIVE dataset was adapted to include HTML-like tags, replacing the default annotations to better suit our named entity translation framework. During fine-tuning of BigTranslate on the MASSIVE dataset, we used loss on validation split from MASSIVE as the criterion for choosing a proper checkpoint.

\subsection{Model Training}
The BigTranslate LLM was fine-tuned using the LoRA technique~\cite{hu2021lora}, as it has been proven effective in adapting LLMs to specific tasks with minimal computational overhead. This fine-tuning process was applied across the MASSIVE dataset for one epoch, emphasizing the translation of named entity slots within the context of the seven targeted languages: German, Spanish, French, Turkish, Hindi, Japanese, Portuguese, and Chinese.

Following the fine-tuning of BigTranslate, the next phase involved translating the English train set from the MultiATIS++ dataset, which contains annotated slots, into the eight languages mentioned. This translation aimed to enrich the dataset for SLU model training, incorporating both the original English data and the newly translated target language data. To mitigate the risk of data leakage and simulate a scenario of limited validation data, the best model was selected based on its performance on a sample English-only trainset from MultiATIS++. Notably, the corresponding translated data were excluded from the training set to prevent any potential bias.

In addition, during the translation process of the MultiATIS++ data into the eight target languages, we implemented a feedback loop. This iterative loop aimed to regenerate the translation multiple times to ensure consistency in the count of HTML-like tags between the source and target sentences.

\subsection{SLU Model Training and Evaluation}

The SLU model training utilized the enriched dataset, comprising original English data and its translated counterparts, to train an SLU model capable of understanding and processing both source and target language inputs. The MultiATIS++ dataset contains 18 intents and 84 slots for each language. The effectiveness of this approach was tested against the official MultiATIS++ test set for each target language, serving as a robust measure of the model's performance and its ability to generalize across languages. 

Our experiments showed significant improvement over the current state-of-the-art method (FC-MTLF), particularly for languages 
considered low-resource in the context of MT and SLU (e.g., Hindi). This improvement is evidence of the effectiveness of named entity slot translation in enhancing multilingual SLU model performance.

While examining the underperformance for Portuguese, we identified that the discrepancies arose from mismatches between the original MultiATIS++ labels and those translated by the BigTranslate model. Notably, the translation model tended to include prepositions (e.g., `a' or `da') within named entities. It was a habit acquired from processing a vast number of examples in the MASSIVE dataset where such prepositions were often part of the slot values, diverging from MultiATIS++'s convention of excluding such prepositions from slot values. As a result, the SLU model, trained on these translated datasets, incorrectly learned to treat these prepositions as integral to slot values. Consequently, when evaluated against the MultiATIS++ test set, which follows the original labeling convention, the model's performance dropped for the slots impacted by this issue. 

\section{Conclusions}

In this article, we showed that state-of-the-art LLM-based MT models are currently the most effective approach to expanding an SLU to new languages.
The experiments validated the proposed approach's effectiveness, surpassing the state-of-the-art performance on the MultiATIS++ dataset. Integration of LLMs with the EasyProject approach showcases a robust and effective solution, advancing the state-of-the-art in cross-lingual SLU. The research assesses the quality of translated data, successfully training an on-device model with a notable \SI{17}{\percent} relative improvement in Overall Accuracy on the MultiATIS++ dataset.

In our study, we examined our method's effectiveness in both cloud-based and on-device SLU scenarios, finding clear differences in their results. Our approach showed a more significant improvement in on-device solutions. This improved performance is directly linked to the differences in model architectures between the two scenarios. Although both models were designed for multilingual use, the inherent limitations of the on-device model in handling multiple languages meant that our method had a more substantial impact.
This outcome highlights the importance of model size in multilingual SLU tasks, especially in environments where efficient resource use is critical.

Finally, our approach demonstrates the effective use of downstream task output to refine LLM-based translations without intensive human effort, streamlining the expansion of SLU to multiple languages. While this work does not employ a Reinforcement Learning Human Feedback Loop (RLHF), including such a feature could provide ways to use feedback from users and experts to improve our methods. This addition could help address the nuanced complexities of language and intent understanding that our current automated feedback loop does not capture, providing a rich ground for future exploration.

\bibliographystyle{IEEEtran}
\bibliography{mybib}

\end{document}